\RequirePackage{fix-cm}

\documentclass[smallextended]{svjour3}       % onecolumn (second format)
\smartqed  % flush right qed marks, e.g. at end of proof
\usepackage{graphicx}
\usepackage{color}
\usepackage{threeparttable}
\usepackage{subfigure}
\usepackage{graphicx}
\usepackage{amsmath}
\usepackage{amssymb}
\usepackage{stfloats}
\usepackage{float}
\usepackage{lipsum}
\usepackage{multirow}
\usepackage[pdftex,colorlinks,bookmarksopen,bookmarksnumbered,citecolor=red,urlcolor=red]{hyperref}
\usepackage{tikz}
\usetikzlibrary{shapes.geometric, arrows, positioning}

\begin{document}

\title{Building Damage Annotation on Post-Hurricane Satellite Imagery Based on Convolutional Neural Networks}

\titlerunning{Building Damage Annotation on Post-Hurricane Satellite Imagery}

\author{Quoc Dung Cao \and 
        Youngjun~Choe}
        
\institute{Quoc Dung~Cao \at
              Department of Industrial and Systems Engineering, University of Washington, Seattle, WA, USA.\\
              Tel.: +1 (217) 979-0117\\
              \email{qcao10@uw.edu}            %  \\
%             \emph{Present address:} of F. Author  %  if needed
           \and
           Youngjun~Choe \at
            Department of Industrial and Systems Engineering, University of Washington, Seattle, WA, USA.\\
             3900 E Stevens Way NE,
Seattle, WA 98195, USA \\
              Tel.: +1 (206) 543-1427\\
              \email{ychoe@u.washington.edu}  
}

\date{Received: date / Accepted: date}

% The correct dates will be entered by the editor

% \thanks{$^{*}$Corresponding author (Postal mailing address: 3900 E Stevens Way NE,
% Seattle, WA 98195, USA; Telephone: +1 (206) 543-1427; FAX: +1 (206) 685-3072; E-mail address: ychoe@u.washington.edu).}
% \thanks{Quoc Dung~Cao and Youngjun~Choe are with the Department of Industrial and Systems Engineering, University of Washington, Seattle, WA, USA.}}% <-this % stops a space
%\thanks{Manuscript received April 19, 2005; revised August 26, 2015.}}

% make the title area
\maketitle

% As a general rule, do not put math, special symbols or citations
% in the abstract or keywords.
%Natural Hazards: Please provide an abstract of 150 to 250 words. The abstract should not contain any undefined abbreviations or unspecified references.
%https://www.springer.com/earth+sciences+and+geography/natural+hazards/journal/11069?detailsPage=pltci_1884672
\begin{abstract}
After a hurricane, damage assessment is critical to emergency managers for efficient response and resource allocation. One way to gauge the damage extent is to quantify the number of flooded/damaged buildings, which is traditionally done by ground survey. This process can be labor-intensive and time-consuming. In this paper, we propose to improve the efficiency of building damage assessment by applying image classification algorithms to post-hurricane satellite imagery. At the known building coordinates (available from public data), we extract square-sized images from the satellite imagery to create training, validation, and test datasets. Each square-sized image contains a building to be classified as either `Flooded/Damaged' (labeled by volunteers in a crowd-sourcing project) or `Undamaged'. We design and train a convolutional neural network from scratch and compare it with an existing neural network used widely for common object classification. We demonstrate the promise of our damage annotation model (over 97\% accuracy) in the case study of building damage assessment in the Greater Houston area affected by 2017 Hurricane Harvey.
\end{abstract}

% Note that keywords are not normally used for peerreview papers.
 %disaster, image classification, neural network, first response, emergency management

\section{Introduction}
When a hurricane makes landfall, situational awareness is one of the most critical needs that emergency managers face before they can respond to the event. To assess the situation and damage, the current practice largely relies on emergency response crews and volunteers to drive around the affected area, which is also known as windshield survey. Another way to assess hurricane damage level is flood detection through synthetic aperture radar (SAR) images (e.g., see the work at the Darthmouth Flood Observatory \cite{dfo}), or the damage proxy map to identify regional-level damages on the built environment (e.g., the Advanced Rapid Imaging and Analysis (ARIA) Project by Caltech and NASA \cite{aria}). SAR imagery is useful in terms of mapping different surface features, texture, or roughness pattern but is harder for laymen to interpret than optical sensor imagery. {\color{black}{The resolutions of virtually all SAR images of today are too coarse to permit the building-level (as opposed to regional-level) damage assessment. Also,  satellites equipped with SAR sensors are far fewer than those with optical sensors, making timely and frequent data collection challenging.}} %It is also harder to collect SAR imagery at a large scale due to the requirement of specialized equipment. Hence, 
In this paper, we focus on using optical sensor imagery as a more intuitive {\color{black}and accessible} way to analyze hurricane damage by distinguishing damaged buildings from the ones still intact. From here onwards, we will refer to optical sensor imagery as `imagery'.

Recently, imagery taken from drones and satellites started to help improve situational awareness from a bird's eye view, but the process still relies on human visual inspection of captured imagery, which is generally time-consuming and unreliable during an evolving disaster. Computer vision techniques, therefore, can be particularly useful. Given the available imagery, our proposed method can automatically annotate \textit{`Flooded/Damaged Building' vs. `Undamaged Building'} on satellite imagery of an area affected by a hurricane. The annotation results can enable stakeholders (e.g., emergency managers) to better plan for and allocate necessary resources. With decent accuracy and quick runtime, this automated annotation process has potential to significantly reduce the time for building situational awareness and responding to hurricane-induced emergencies. 

\begin{figure}[ht!]
{\centering
\includegraphics[height=3in]{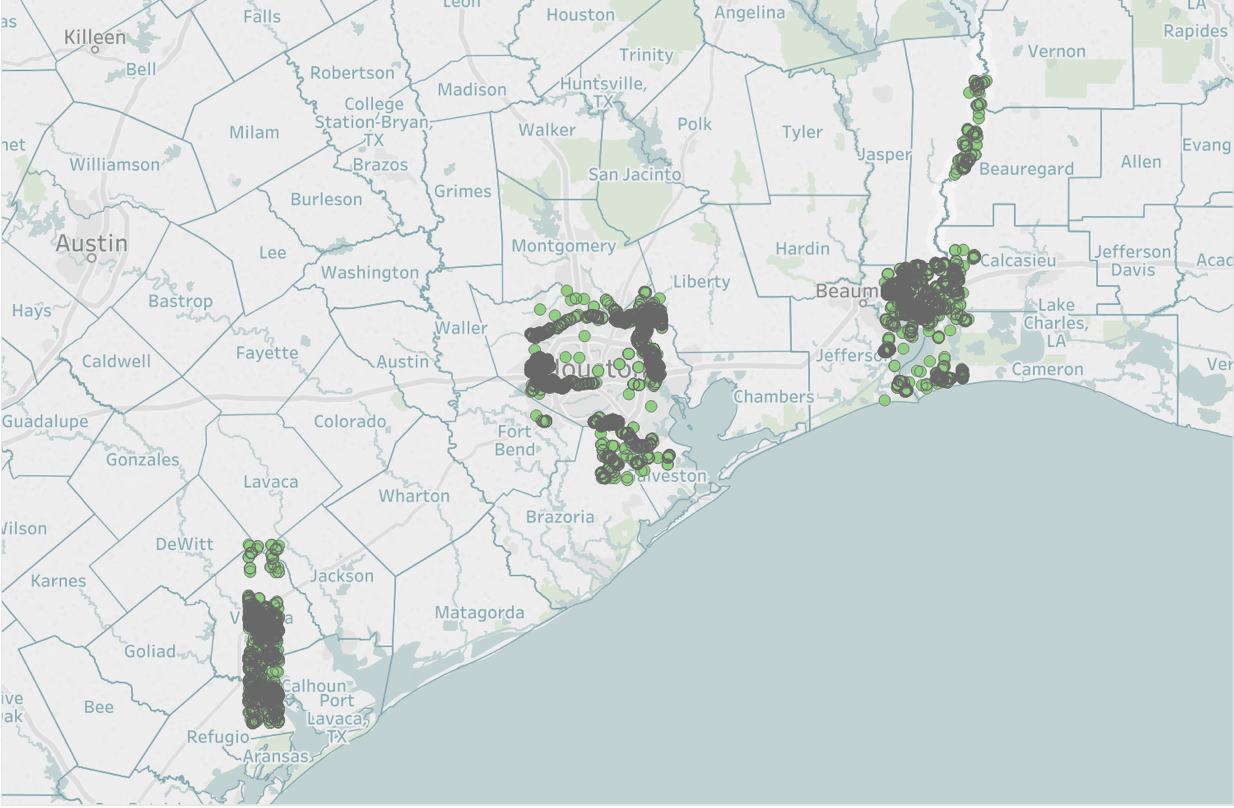} \\
\centering\caption{\small{The Greater Houston area was affected by Hurricane Harvey in 2017. The green circles represent the coordinates of flooded/damaged structures tagged by Tomnod volunteers.}}
\label{fig:Harvey}
}
\end{figure}

The satellite imagery data used in this paper covers the Greater Houston area before and after Hurricane Harvey in 2017 (Figure~\ref{fig:Harvey}). The flooded/damaged buildings were labeled by volunteers through the crowd-sourcing project, Tomnod \cite{tomnod}. We then process, filter, and clean the dataset to ensure that it has correct labels and can be learned appropriately by a learning algorithm. 

% The satellite imagery data used in the paper was captured by optical sensors with sub-meter resolution and preprocessed for orthorectification, atmospheric compensation, and pansharpening from the Greater Houston area before and after Hurricane Harvey in 2017 (Figure~\ref{fig:Harvey}). The damaged buildings were labeled by volunteers through the crowd-sourcing project Tomnod \cite{tomnod}. We then process, filter, and clean the dataset to ensure that it has higher quality and can be learned appropriately by the deep learning algorithm. 

% below is for submission only, use above for publication
% \begin{figure}[h]
% {\centering
% \includegraphics[height=2.5in]{images/Harvey2017.png} \\
% \centering\caption{\small{Affected areas during Hurricane Harvey in 2017. The green dots are coordinates with damaged buildings/roads tagged by volunteers.}}
% \label{fig:Harvey}
% }
% \end{figure}

By sharing the dataset and code used in this paper (see the appendix), we hope that other researchers can build upon this study and help further improve computer vision-based damage assessment process. %use the dataset to study and experiment with different uses of satellite imagery in disaster response. In addition, our method can be applied to future hurricane events to improve damage assessment and resource planning. 
The shared code includes a pre-trained deep-learning architecture that achieves %satisfactory result in terms of 
the best classification accuracy (detailed in Section~\ref{sec:case_study}). It can facilitate transfer learning either in feature extraction, fine-tuning, or as a baseline model to speed up the learning process for future hurricane events. % with similar properties. 

The remaining of this paper is organized as follows. In Section~\ref{sec:background}, we present a brief review of convolutional neural networks, machine learning-based damage annotation work on post-hurricane satellite imagery, and challenges in the damage annotation on satellite imagery. Section~\ref{sec:method} describes our proposed methodological framework for the damage annotation. Details of the implementation and discussion of the results are presented in Section~\ref{sec:case_study}. Finally, Section~\ref{sec:conclusion} concludes this paper and draws some future research directions.

\section{Background}\label{sec:background}
\subsection{Convolutional neural network}
%Object detection is a ubiquitous topic in computer vision, thanks to the development of 
The convolutional neural network (CNN) \cite{cnn} often yields outstanding results over other algorithms for computer vision tasks such as %natural language processing \cite{cnn-nlp}, 
object categorization \cite{cnn-object}, image classification \cite{cnn-image,cnn-imagenet}, and object recognition \cite{trafficSign}. Variations of CNN have been successfully applied to remote sensing image processing tasks \cite{Zhang2016} such as aerial scene classification \cite{aid,aerial-label,scene-multiscale}, SAR imagery classification \cite{sar-cnn}, or object detection in unmanned aerial vehicle imagery \cite{Bazi2018}.

\begin{figure*}[ht!]
{\centering
\includegraphics[width=5in]{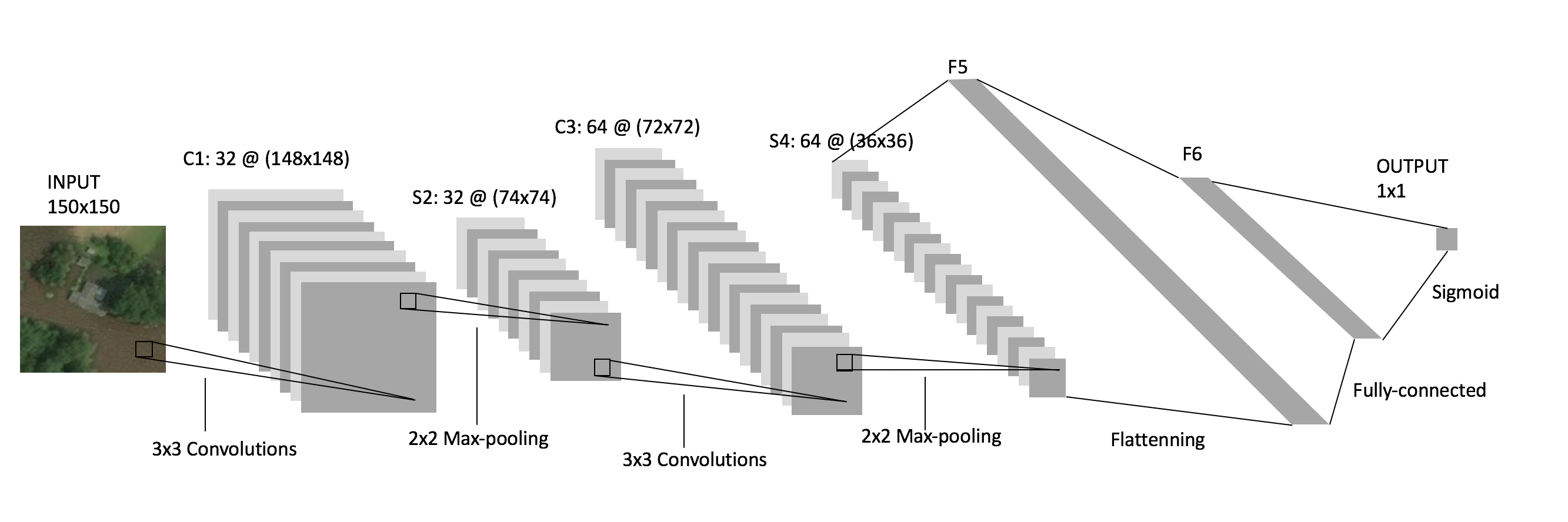} \\
\centering\caption{A convolutional neural network inspired by LeNet-5 architecture in \cite{cnn}; C: Convolutional layer, S: Sub-sampling layer, F: Fully connected layer; 32@(148x148) means there are 32 filters to extract features from the input image, and the original input size of 150x150 is reduced to 148x148 since no padding is added around the edges during $3\times3$ convolution operations so 2 edge rows and 2 edge columns are lost; 2x2 Max-pooling means the data will be reduced by a factor of 4 after each operation; Output layer has 1 neuron since the network outputs the probability of one class (`Flooded/Damaged Building') for binary classification.}% , and the other probability is just 1-$\mathbb{P}$r(`Flooded/Damaged Building').}
\label{fig:cnn-archi}
}
\end{figure*}

Structurally, CNN is a feed-forward network that is particularly powerful in extracting hierarchical features from images. The common structure of CNN has three components: the convolutional layer, the sub-sampling layer, and the fully connected layer as illustrated in Figure~\ref{fig:cnn-archi}. 

In the convolutional layer (C in Figure~\ref{fig:cnn-archi}), each element (or neuron) of the network in a layer receives information from a small region of the previous layer. A 3x3 convolutional filter will take a dot product of 9 weight parameters with 9 pixels (3x3 patch) of the input, and the resulting value is transformed by an activation function to become a neuron value in the next layer. The same region can yield many information maps to the next layer through many convolutional filters. In Figure~\ref{fig:cnn-archi}, at convolutional layer C1, we have 32 filters that represent 32 ways to extract features from the previous layers and form a stack of 32 feature matrices. Another advantage of CNN is its robustness to shift of features in the input images \cite{shift-invariant}. This is crucial since in many datasets, objects of interest are not necessarily positioned right at the center of the images and we want to learn the features, not their positions. 

In the sub-sampling layer (S in Figure~\ref{fig:cnn-archi}), the network performs either local averaging or max pooling over a patch of the input. If the sub-sampling layer size is 2x2 such as S2, local averaging will yield the mean of the 4 nearby convoluted pixel values, whereas max pooling will yield the maximum value among them. Essentially, this sub-sampling operation reduces the input feature matrix to half its number of columns and rows, which helps to reduce the resolution by a factor of 4 and the network's sensitivity to distortion. 

After the features are extracted and the resolution reduced, the network will flatten the final stack of feature matrices into a feature vector and pass it through a sequence of fully connected layers (F in Figure~\ref{fig:cnn-archi}). Each subsequent layer's output neuron is a dot product between the feature vector and a weight vector, transformed by a non-linear activation function. In this paper, the last layer has only 1 neuron, which is the probability of a reference class (`Flooded/Damaged building'). 

As mentioned, the dot products are transformed by an activation function. This gives a neural network, with adequate size, the ability to model any function. Some common activation functions include sigmoid $f(x) = \frac{1}{1+e^{-x}}$, rectified linear unit (ReLU) $f(x) = max(0, x)$, and leaky ReLu $f(x) = max(\alpha x, x)$, with $0 < \alpha \ll 1$. There is no clear reason to choose any specific function over the others to improve performance of a network. However, using ReLU may speed up the training of the network without affecting the performance \cite{relu}.

\subsection{Machine learning-based damage annotation on post-hurricane satellite imagery}
% Within the area of remote sensing, object detection on satellite imagery is a well-established research area. Variations of CNNs and other approaches were applied in different tasks and dataset, although the existing studies focus on detecting roads, buildings, trees, vehicles, ships, airplanes, or airports \cite{Cheng2016,Zhang2016}. Recurrent neural network \cite{Zhu2017} was used to classify different infrastructure types and agriculture crops. Xia et al. constructed the Aerial Image Dataset (AID), a large-scale dataset so that other researchers can benchmark their scene classification algorithms \cite{aid}. CNN and support vector machine were also applied to to unmanned aerial vehicle imagery, e.g. captured by drone \cite{Bazi2018}, in which a custom CNN was shown to outperform a pre-trained network. 

%Within the area of remote sensing, classification on satellite imagery is a well-established research area. Nevertheless,
% Machine learning on remote sensing imagery  is actively researched to assess damage from or susceptibility to various hazards such as earthquake \cite{Ranjbar2018}, landslide \cite{Ada2018,Hong2018}, tsunami \cite{Mehrotra2015}, and wildfire \cite{Lu2018}. Such methods showed remarkable promise, but leveraging unique characteristics of each hazard, they are not directly applicable to damage annotation on post-hurricane imagery.

Machine learning on remote sensing imagery  is actively researched to assess damage from or susceptibility to various hazards such as earthquake \cite{Ranjbar2018}, landslide \cite{Ada2018,Hong2018}, tsunami \cite{Mehrotra2015}, and wildfire \cite{Lu2018}. {\color{black}Such methods showed remarkable promise. However, often leveraging unique characteristics of each hazard type, they are not directly applicable to damage annotation on post-hurricane imagery and sometimes require extensive pre-processing. For example, the work in \cite{Ranjbar2018} for post-earthquake damage assessment closely resembles our work. Both pre-event and post-event are used to extract building's roof and texture features. This process requires input from expert operators and rely on texture feature of the buildings in the imagery, which sometimes may not be available depending on means of collection. In another work, \cite{Mehrotra2015} classifies regions of pixels into water, vegetation, urban, and bare land, which does not provide the building level granularity as we pursue. Furthermore, since we rely purely on the widely available optical sensor imagery, convolutional neural network is one of the most suitable model classes due to its flexible feature extraction capability and architecture.}

Some recent studies used machine learning to assess post-hurricane damages on satellite imagery. A small project studied detecting \textit{flooded roads} by comparing pre-event and post-event satellite imagery \cite{Jack2017} but the method is not applicable to other types of damages. Two commercial vendors of satellite imagery also separately developed unsupervised algorithms to detect flooded area using spectral signature of impure water (which is not available from the pansharpened satellite images in our data) \cite{planet,gbd}. Before deep learning era, a method using a pattern recognition template set was applied to detect hurricane damages in \textit{multispectral} images \cite{Barnes2007} but the method is not applicable to our pansharpened images. 

\subsection{Challenges in damage annotation on satellite imagery}
\begin{figure}[ht!]
{\centering
\subfigure[Lower-quality orthorectification]{\includegraphics[width=2in]{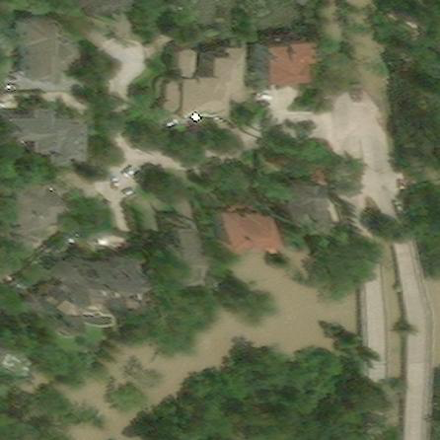}}
\subfigure[Higher-quality orthorectification]{\includegraphics[width=2in]{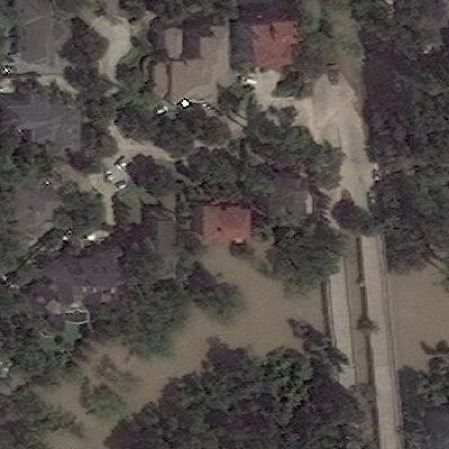}}
\subfigure[More blurry]{\includegraphics[width=2in]{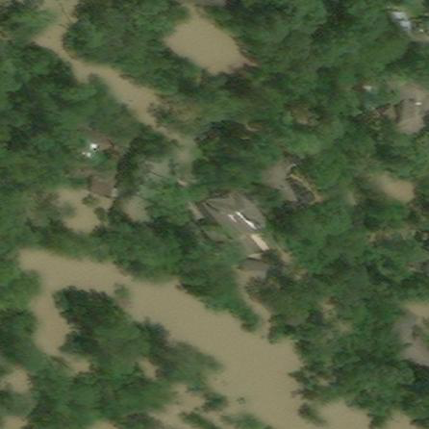}}
\subfigure[Less blurry]{\includegraphics[width=2in]{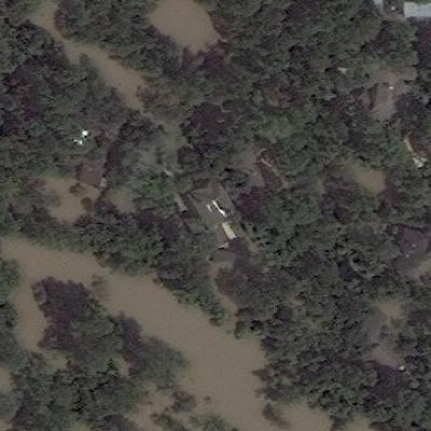}}
\\
\caption{\small{Different orthorectification and pre-processing quality of the same location on different days.}}
\label{fig:orthorectification}
}
\end{figure}

There are multiple challenges in damage annotation on satellite imagery. First, satellite imagery resolution is not as high as various benchmark datasets commonly used to train neural networks (NNs) (e.g., ImageNet \cite{cnn-imagenet} and traffic signs \cite{trafficSign}) with respect to the objects of interest. Dodge \& Karam \cite{Dodge2016} studied the performance of NNs under quality distortions and highlighted that NNs could be prone to errors in blurry and noisy images. Although our dataset is of relatively high resolution (e.g., one of the satellites capturing the imagery is GeoEye-1, which has 46cm panchromatic resolution \cite{geoeye}), it is still far from the resolution of common-object detection datasets (e.g., animals, vehicles). In fact, the labeling task on satellite imagery is hard even with human visual inspection, which leads to another challenge. The volunteers' annotation could be erroneous. To limit this, the crowd-sourcing platform has a proprietary system that computes the agreement score of each label. In this paper, we ignore this information to gather as many labels as possible and take the given labels as ground truth since limited size of training data could be a critical bottle-neck for models with many parameters to learn such as NNs. Third, there are some inconsistencies in image quality. Since the same region can be captured multiple times on different days, the same coordinate may have multiple images of different qualities (e.g., due to pre-processing), as shown in Figure~\ref{fig:orthorectification}. In summary, effective learning algorithms should overcome the challenges from low-resolution images, noisy labels, and inconsistent image qualities. 

% \vspace{-.3cm}
%%%%%%%%%%%%%%%%%%%%%%%%%%%%%%%%%%%%%%%%%%%%%%%%%%%%%%%%%%%%%%%%%%%%%%%%%%%%%%
% Methodology
%%%%%%%%%%%%%%%%%%%%%%%%%%%%%%%%%%%%%%%%%%%%%%%%%%%%%%%%%%%%%%%%%%%%%%%%%%%%%%
\section{Methodology}\label{sec:method}
In this section, we describe our end-to-end methodological framework from collecting, processing, featurizing data to building the convolutional neural network to classify whether a building in a satellite image is flooded/damaged  or not. 

\subsection{Data description}
\begin{figure}[ht!]
{\centering
\includegraphics[height=3.2in]{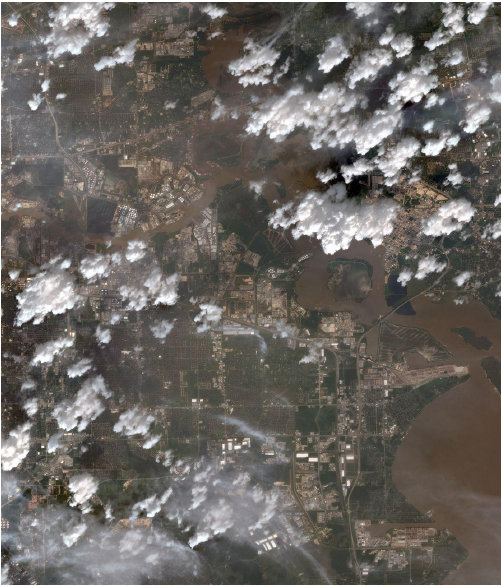} \\
\centering\caption{\small{A typical strip of image in the dataset.}}
\label{fig:strip}
}
\end{figure}

\begin{figure}[ht!]
{\centering
\subfigure[Blacked out partially]{\includegraphics[width=2in]{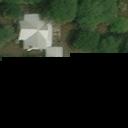}}
\subfigure[Covered by cloud partially]{\includegraphics[width=2in]{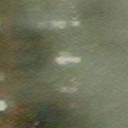}}
\subfigure[Covered by cloud mostly]{\includegraphics[width=2in]{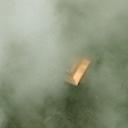}}
\subfigure[Covered by cloud totally]{\includegraphics[width=2in]{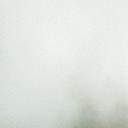}}
\\
\caption{\small{Examples of discarded images during the data cleaning process due to their potential to hinder model training.}}
\label{fig:bad_images}
}
\end{figure}

 The satellite imagery of the Greater Houston area was captured by optical sensors with sub-meter resolution, preprocessed (e.g., orthorectification and atmospheric compensation), and pansharpened by the image provider. The raw imagery consists of around four thousand image strips taken on multiple days (each strip is roughly 1GB and has around 400 million pixels with RGB bands). Some strips overlap and have black pixels in the overlapped region. Some images are also covered fully or partially by clouds. Figure~\ref{fig:strip} shows a typical strip in the dataset and Figure~\ref{fig:bad_images} shows some examples of \textit{low quality} images (from the perspective of model training) that we chose to discard.
% \begin{enumerate}
%     \item The raw imagery data covering the Greater Houston area was captured in about four thousand strips ($\sim$ 400 million pixels ($\sim$ 1GB) with RGB bands per strip) in different days. Hence, some strips can overlap, leading to some images blacked out at the boundaries. In some days, the images are also covered fully or partially by clouds.
% Figure~\ref{fig:strip} shows a typical strip in the data set and Figure~\ref{fig:bad_images} shows some examples of low quality images in 128x128 pixels that we choose to discard.

% Due to the big volume, it is not feasible to store the data in a local computer. The raw imagery data was downloaded and stored on a high performance computing cluster.

% \begin{figure}[h]
% {\centering
% \subfigure[Blacked out partially]{\includegraphics[width=1.4in]{bad_images/bad1.jpeg}}
% \subfigure[Covered by cloud partially]{\includegraphics[width=1.4in]{bad_images/bad2.jpeg}}
% \subfigure[Covered by cloud mostly]{\includegraphics[width=1.4in]{bad_images/bad3.jpeg}}
% \subfigure[Covered by cloud totally]{\includegraphics[width=1.4in]{bad_images/bad4.jpeg}}
% \\
% \caption{\small{Examples of 128x128-pixel low quality images}}
% \label{fig:bad_images}
% }
% \end{figure}

%     \item The raw data is in geoTIFF format
% %     , which allows georeference information to be embedded within a TIFF file.
%     \item \textit {Flooded/Damaged} buildings (or \textit{Damaged}) are annotated with labels and coordinates given in GeoJSON format. Using the coordinates, we extract the images of damaged buildings in JPEG format from the geoTIFF \textit{post-event} imagery. 
%     \item \textit{Undamaged buildings} (or \textit{Undamaged}) are extracted in JPEG format directly from the geoTIFF \textit{pre-event} imagery at the same coordinates.
% \end{enumerate}

\subsection{Damage annotation}\label{sec:annotation}
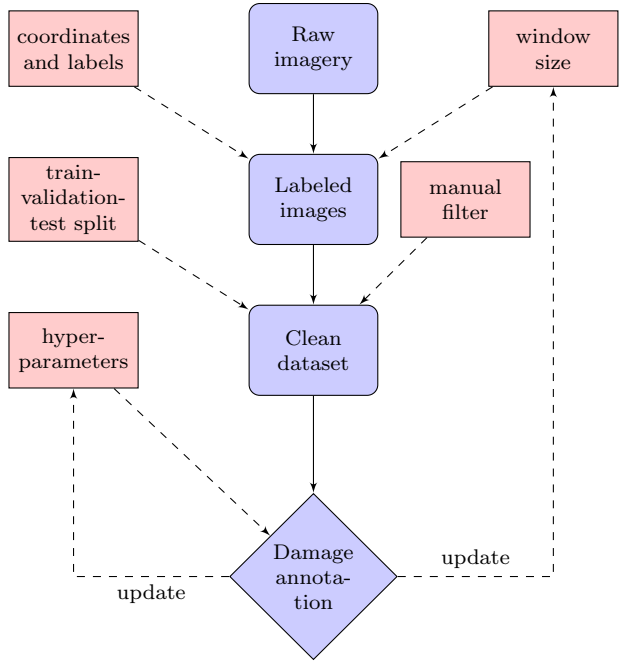
\begin{figure}
\begin{center}
% Define block styles
\tikzstyle{decision} = [diamond, draw, fill=black!20, 
    text width=4.5em, text badly centered, node distance=3cm, inner sep=0pt]
\tikzstyle{block} = [rectangle, draw, fill=black!20, 
    text width=5em, text centered, rounded corners, minimum height=4em]
\tikzstyle{line} = [draw, -latex']
\tikzstyle{cloud} = [draw, rectangle, minimum width=1.5cm, minimum height=1cm, text centered, text width=1.5cm, draw=black, fill=red!20]
    
\begin{tikzpicture}[node distance = 2cm, auto]
    % Place nodes
    \node [block] (input) {Raw imagery};
    \node [cloud, left of=input] (coordinate) [left = 0.3cm]{coordinates and labels};
    \node [cloud, right of=input] (size) [right = 0.3cm]{window size};
    \node [block, below of=input] (label) {Labeled images};
    \node [cloud, left of=label] (split) [left = 0.3cm] {train-validation-test split};
    \node [cloud, right of=label] (filter) {manual filter};
    \node [block, below of=label] (clean) {Clean dataset};
    \node [cloud, left of=clean] (hyper) [left = 0.3cm] {hyper-parameters};
    \node [decision, below of=clean] (decide) {Damage annotation};
    % Draw edges
    \path [line] (input) -- (label);
    \path [line] (label) -- (clean);
    \path [line] (clean) -- (decide);
    \path [line,dashed] (coordinate) -- (label);
    \path [line,dashed] (size) -- (label);
    \path [line,dashed] (split) -- (clean);
    \path [line,dashed] (filter) -- (clean);
    \path [line,dashed] (hyper) -- (decide);
    \path [line,dashed] (decide) -| node [near start] {update} (hyper);
        \path [line,dashed] (decide) -| node [near start] {update} (size);
    %\path [line,dashed] (system) |- (evaluate);
\end{tikzpicture}
\end{center}
\caption{The damage annotation framework.} \label{fig:framework}
\end{figure}

We present here our methodological framework (Figure~\ref{fig:framework}) that starts from raw data input to create damage annotation output. The first step is to process the raw data to create training-ready data by using a cropping window approach. Essentially, the building coordinates, which can be easily obtained from public data (e.g., OpenStreetMap \cite{osm}), can be used as the centers of cropping. We use the building coordinates already associated with the damage labels from Tomnod. A window is then cropped from the raw satellite imagery to create a data sample. Tomnod volunteers' annotation of flooded/damaged buildings is taken as the ground truth for the positive label, `Flooded/Damaged building'. At the same coordinates, we crop windows from the imagery captured before the hurricane to create negative data samples, labeled `Undamaged building'.

The optimal window size depends on various factors including the image resolution and building footprint sizes. Too small windows may limit the background information contained in each sample, whereas too large ones may introduce unnecessary noise. We keep the window size as a tuning hyper-parameter in the model. A few sizes are considered such as 400x400, 128x128, 64x64, and 32x32. 

The cropped images are then manually filtered to ensure the high quality of the dataset. To let the model generalize well, we only discard the images that can obviously hamper the algorithm's learning process, such as the example images in Figure~\ref{fig:bad_images}. The cleaned images are then split into training, validation, and test sets and fed to a convolutional neural network for damage annotation as illustrated in Figure~\ref{fig:framework}. Validation accuracy is monitored to tune the necessary hyper-parameters (including the window size).

% \begin{tikzpicture}[node distance = 2.8cm, auto, cloud/.style={draw,
% fill=pink,
% rectangle, 
% minimum width={width("hyper-paramters")+2pt}}]
%     % Place nodes
%     \node [block] (input) {Raw imagery};
%     \node [cloud, left of=input] (coordinate) [left = 1cm]{coordinates and labels};
%     \node [cloud, right of=input] (size) [right = 1cm]{window size};
%     \node [block, below of=input] (label) {Labeled images};
%     \node [cloud, left of=label] (split) [left = 1cm] {train-validation-test split};
%     \node [cloud, right of=label] (filter) {manual filter};
%     \node [block, below of=label] (clean) {Clean dataset};
%     \node [cloud, left of=clean] (hyper) [left = 1cm] {hyper-parameters};
%     \node [decision, below of=clean] (decide) {Damage annotation};
%     % Draw edges
%     \path [line] (input) -- (label);
%     \path [line] (label) -- (clean);
%     \path [line] (clean) -- (decide);
%     \path [line,dashed] (coordinate) -- (label);
%     \path [line,dashed] (size) -- (label);
%     \path [line,dashed] (split) -- (clean);
%     \path [line,dashed] (filter) -- (clean);
%     \path [line,dashed] (hyper) -- (decide);
%     \path [line,dashed] (decide) -| node [near start] {update} (hyper);
%         \path [line,dashed] (decide) -| node [near start] {update} (size);
%     %\path [line,dashed] (system) |- (evaluate);
% \end{tikzpicture}
% \end{center}

% \caption{Damage annotation framework} \label{fig:framework}

% \end{figure}

\subsection{Data processing}\label{sec:collection}
% \begin{enumerate}
As described above, the data generation starts from a building coordinate. Since there are multiple raw images containing the same coordinates, there are duplicate images with different quality. This can potentially inflate the prediction accuracy as the same coordinate may appear in both the training and test sets. We maintain a set of the available coordinates and make sure each coordinate is associated with a unique, ``good-quality" image in the final dataset through a semi-automated process. We first automatically discard the totally blacked out images for each coordinate, and keep the first image we encounter that is not totally black. The resulting set of images are manually filtered to eliminate the images that are partially black or covered by clouds. 
%     \item Many images are just blank due to the original raw image. We need to discard those and keep a valid image for each coordinate. Pick the first one that is not blank in the list of duplicates
%     \item Manually pick the images of good quality. It takes hours. 
%     \item Manually pick the training, validation, and test set to make sure the representation is good, since they can be close to one another and "almost" repeat themselves. 
% \end{enumerate}

% We manually filter the bad-quality images and remove duplicates to make sure the dataset contains only one good-quality image at every unique coordinate.

\subsection{Data featurization}\label{sec:featurization}
\begin{figure}[ht!]
{\centering
\subfigure[Original image (Flooded/Damaged)]{\includegraphics[width=2in]{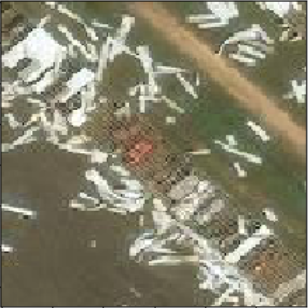}}
\subfigure[After $1^{st}$ layer]{\includegraphics[width=2in]{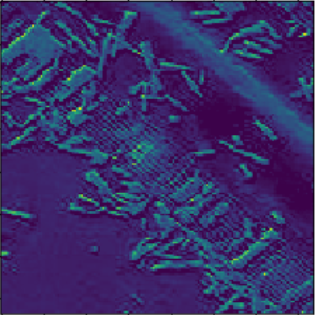}}
\subfigure[After $2^{nd}$ layer]{\includegraphics[width=2in]{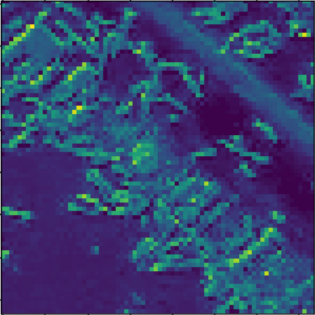}}
\subfigure[After $3^{rd}$ layer]{\includegraphics[width=2in]{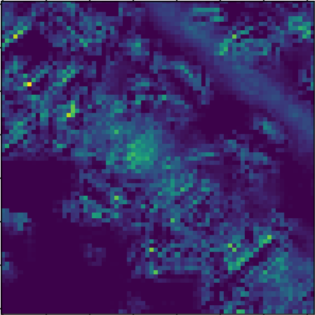}}
\\
\caption{\small{Information flow within one filter after each convolutional layer. The initial layers act as a collection of edge extraction. At a deeper layer, the information is more abstract and less visually intepretable.}}
\label{fig:one_filter}
}
\end{figure}

Since we control the window size based on physical distance, there could be round-off errors when converting the distance to the number of pixels. Therefore, we project them into the same feature dimension. For instance, both a 128x128 image and a 127x129 image are projected into 150x150 dimension. The images are then fed through a CNN to further extract useful features, such as edges, as illustrated in Figure~\ref{fig:one_filter}.

How to construct the most suitable CNN architecture is an ongoing research problem. The common practice, known as \textit{transfer learning}, is starting with a  known architecture and fine-tuning it. We experiment with a well-known architecture, VGG-16 \cite{vgg16}, and modify the first layer to suit our input dimension. VGG-16 is known to perform very well on the ImageNet dataset for common object classification. 

\begin{figure}[ht!]
{\centering
\subfigure[After $1^{st}$ layer]{\includegraphics[width=2.2in]{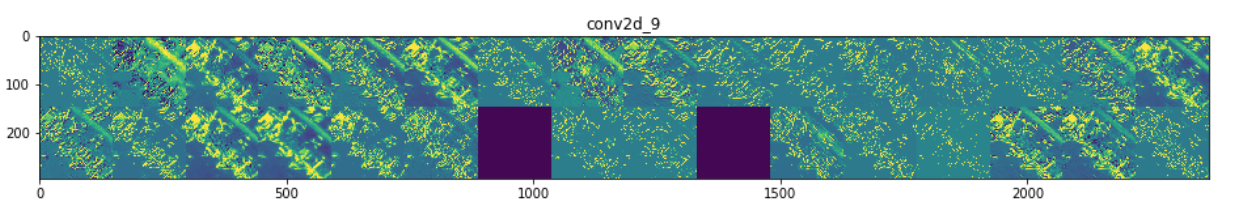}}
\subfigure[After $2^{nd}$ layer]{\includegraphics[width=2.2in]{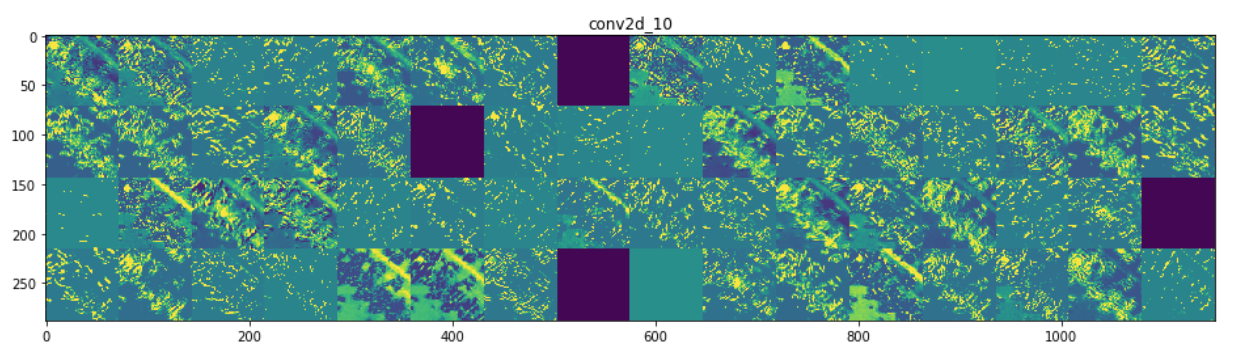}}
\subfigure[After $3^{rd}$ layer]{\includegraphics[width=2.2in]{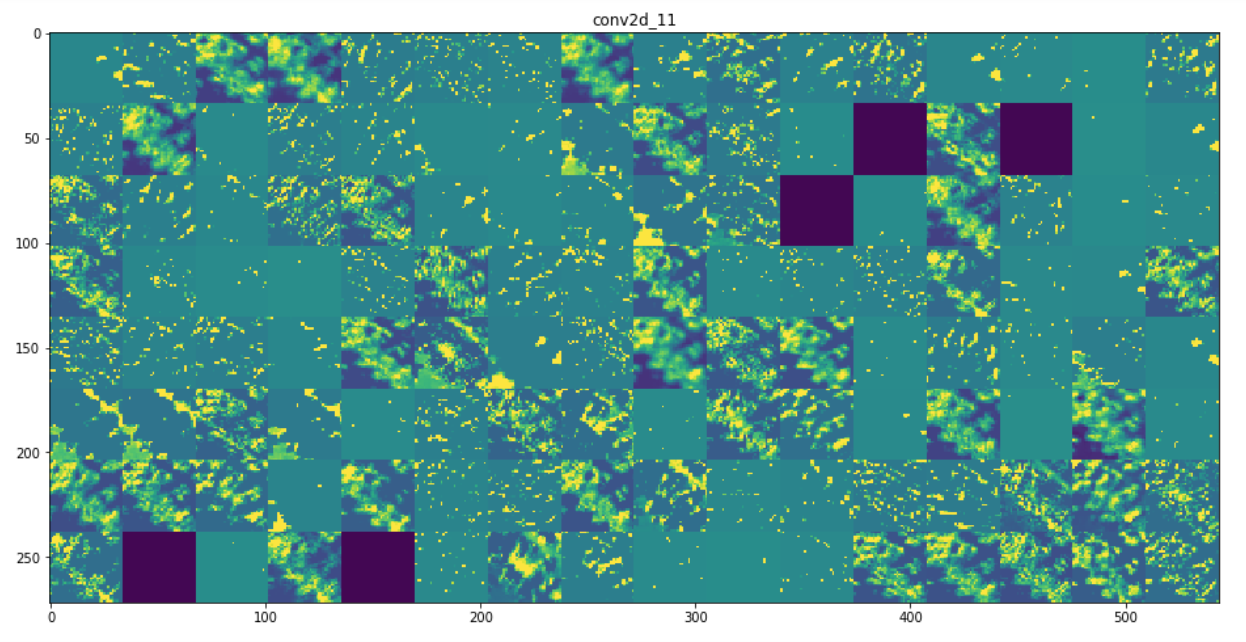}}
\subfigure[After $4^{th}$ layer]{\includegraphics[width=2.2in]{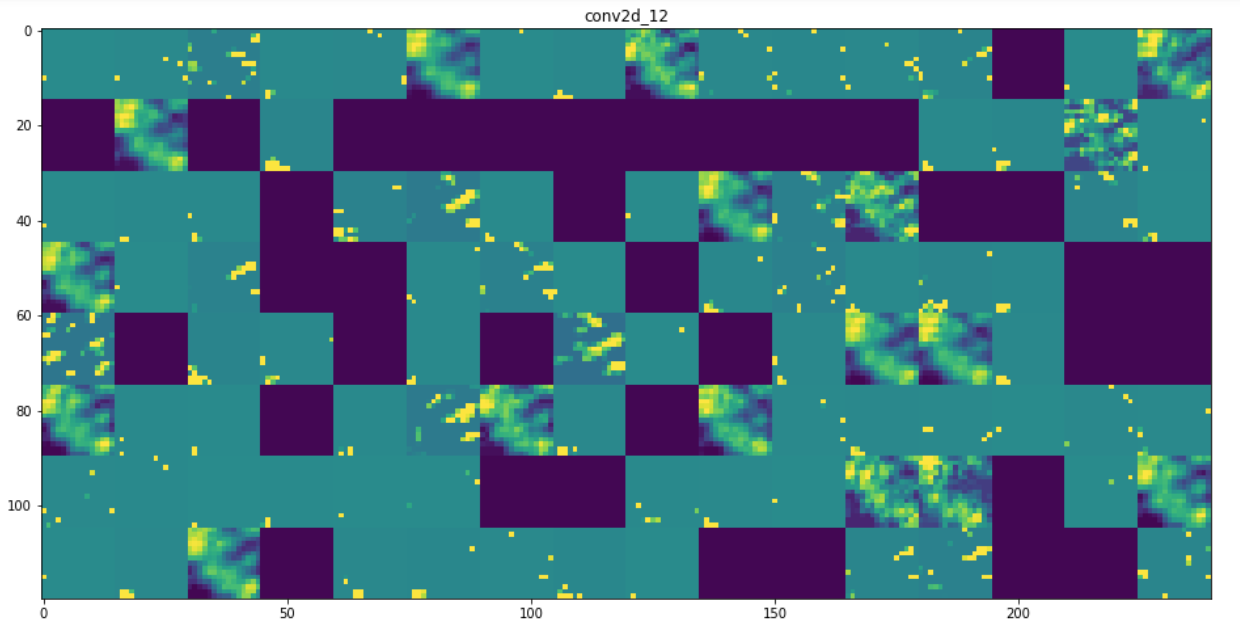}}
\\
\caption{\small{Information flow in all filters after each convolutional layer. The sparsity increases with the depth of the layer, as indicated by the increasing number of dead filters.}}
\label{fig:flow}
}
\end{figure}

However, because of the substantial differences between the common object classification and our flooded/damaged building classification, we also build our own network from scratch. We carefully consider proper hyper-parameters, as similarly done in \cite{customized-cnn}. Our basis for determining the size and depth of a customized network is to monitor the information flow through the network and stop enlarging the network when there are too many \textit{dead} filters (\textit{i.e.}, blank filters that do not carry any further information to the subsequent layers in the network). Due to the nature of the rectified linear unit (ReLU), which is defined as $\max(0,x)$, there will be many zero weights in the hidden layers. Although sparsity in the layers can promote the model to generalize better, it may cause the problem on gradient computation at 0, which in turns does not update any parameters, and hurt the overall model performance \cite{relu,leaky}. We see that in Figure ~\ref{fig:flow} after four convolutional layers, about 30\% of the filters are dead and will not be activated further. This is a significant stopping criterion since we can avoid a deep network such as VGG-16 to save the computational time and safeguard satisfactory information flow in the network at the same time. 

We present our customized network architecture that achieves the best result in Table \ref{tab:architecture}. The network begins with four convolutional and max pooling layers and ends with two fully connected layers. 

% An example of how the number of parameters is calculated is as follows: in the first 2-D convolutional layer, there will be 32 convolutional filters, each of size (3x3), for each of the 3 RGB channels of the input image. Including 1 more bias parameter for each filter, we have: \\
% \[
% [(3 \times 3) \times 3 + 1] \times 32 = 896
% \]
In our CNN structure, with four convolutional layers and two fully connected layers, there are already about $3.5$ million parameters to train, given $67,500$ pixels as an input vector for each image. The VGG-16 structure \cite{vgg16}, with thirteen convolutional layers, has almost $15$ million trainable parameters, which can over-fit, require more resources, and reduce generalization performance on the testing data. In addition, as discussed in \cite{customized-cnn}, the network depth should depend on the complexity of the features to be extracted from the image. Since we have only two classes of interest, a shallower network can be favourable in terms of training time and generalization.

\begin{table}[t]
\caption{Convolutional neural network architecture that achieves the best result.}
\begin{center}
\begin{tabular}{lllll}
\multicolumn{1}{c}{\bf Layer type}  &\multicolumn{1}{c}{\bf Output shape}
   &\multicolumn{1}{c}{\bf \begin{tabular}{@{}c@{}}Number of \\ trainable parameters \end{tabular}}

\\ \hline 
Input                       &3@(150x150)     &0  \\
2-D Convolutional 32@(3x3)        &32@(148x148)       &896   \\
2-D Max pooling (2x2)        &32@(74x74)       &0   \\
2-D Convolutional 64@(3x3)        &64@(72x72)       &18,496  \\
2-D Max pooling (2x2)         &64@(36x36)     &0   \\
2-D Convolutional 128@(3x3)        &128@(34x34)       &73,856   \\
2-D Max pooling (2x2)        &128@(17x17)       &0   \\
2-D Convolutional 128@(3x3)        &128@(15x15)       &147,584   \\
2-D Max pooling (2x2)        &128@(7x7)       &0   \\
Flattening      &1x6272       &0  \\
Dropout         &1x6272       &0  \\
Fully connected layer &1x512       &3,211,776  \\
Fully connected layer &1x1       &513  \\
\hline
\end{tabular}
\end{center}
\textit{Note}: The total number of trainable parameters is 3,453,121.
C@($A\times B$) is interpreted as that there are a total of C matrices of shape ($A\times B$) stacked on top of one another to form a three-dimensional tensor.
2-D Max pooling layer with ($2\times 2$) pooling size means that the input tensor's size will be reduced by a factor of 4. \\
\label{tab:architecture}
\end{table}

% \begin{figure}[h]
% {\centering
% \subfigure[After $1^{st}$ layer]{\includegraphics[width=2.5in]{images/flow1.png}}
% \subfigure[After $2^{nd}$ layer]{\includegraphics[width=2.5in]{images/flow2.png}}
% \subfigure[After $3^{rd}$ layer]{\includegraphics[width=2.5in]{images/flow3.png}}
% \subfigure[After $4^{th}$ layer]{\includegraphics[width=2.5in]{images/flow4.png}}
% \\
% \caption{\small{Information flow in all filters after each layer}}
% \label{fig:flow}
% }
% \end{figure}

% Our network is shallower but can outperform the pre-built network. 

% \begin{enumerate}
%     \item Project to feature space of unequal image sizes
%     \item Convolutional + Max pooling hyper parameter
%     \item Unfeaturized image trial if time allows
% \end{enumerate}

\subsection{Image classification}\label{sec:classification}
Due to the limited availability of pre-event images and the exclusion of some images (e.g., due to cloud coverage) in the \textit{Flooded/Damaged} and \textit{Undamaged} categories, our dataset is unbalanced with the majority class being \textit{Flooded/Damaged}. Thus, we split the dataset into training, validation, and test datasets as follows. We keep the training and validation sets balanced and leave the remaining data to construct two test sets, a balanced set and an unbalanced (with a ratio of 1:8) set. 

The first performance metric is the classification accuracy. In contrast to the balanced test set, we note that the baseline accuracy for the \textit{unbalanced} test set  is ${8}/{9}=88.89\%$ (greater than the random guess accuracy, 50\%), which can be achieved by annotating all buildings as the majority class \textit{Flooded/Damaged}. In addition, as the classification accuracy is sometimes not the most pertinent performance measure, we also monitor the area under the receiver operating characteristic curve (AUC), which is a widely-used criterion to measure the classification ability of a binary classifier under a varying decision threshold \cite{auc}.

% Leaky ReLu may be better than ReLu for satellite images.

% Optimizer: Adam \cite{Adam} and RMSprop
% \begin{enumerate}
%     \item Benchmark accuracy: predict all to be the majority class
%     \item Unbalance class problem. Balance training data, plus hard negative mining, unbalanced validation and test
%     \item Logistic Regression featurize all the images first and then fit Logistic Reg
%     \item MLP (monitor gradient vanishing and information flow through layer to set number of layers and activation function, initialization): MLP with ReLu, with Leaky Relu, with Data Augmentation. 
%     \item Sensitive analysis on accuracy to build confidence interval of cross-validation errors
%     \item Show the dead neuron or some image flows through the layers + loss and accuracy curve

%     \item Show some images of misclassification and make some sense out of it
% \end{enumerate}

%%%%%%%%%%%%%%%%%%%%%%%%%%%%%%%%%%%%%%%%%%%%%%%%%%%%%%%%%%%%%%%%%%%%%%%%%%%%%%
% Case Study
%%%%%%%%%%%%%%%%%%%%%%%%%%%%%%%%%%%%%%%%%%%%%%%%%%%%%%%%%%%%%%%%%%%%%%%%%%%%%%
\section{Implementation and Result}\label{sec:case_study}
We train the neural networks using the \textit{Keras} library with TensorFlow backend with a single NVIDIA K80 Tesla GPU. The network weights are initialized using Xavier initializer \cite{xavier}. The mini batch size for the stochastic gradient descent optimizer is 32.

After the data cleaning process, our dataset contains 14,284 positive samples (\textit{Flooded/Damaged}) and 7,209 negative samples (\textit{Undamaged}) at unique geographical coordinates. 5,000 samples of each class are in the training set. 1,000 samples of each class are in the validation set. The rest of the data are reserved to form the test sets, i.e. in the balanced test set, there will be 1,000 samples of each class, and in the unbalanced test set, there will be 8,000 samples of \textit{Flooded/Damaged} class and  1,000 samples of \textit{Undamaged} class. 

Due the expensive computational cost of training the CNN, we investigate selected combinations of the hyper-parameters in a greedy manner, instead of tuning all the hyper-parameters through a full grid search or full cross-validation. For example, we investigate the performance of a model with multiple window sizes (400x400, 128x128, 64x64, and 32x32) and select the 128x128 window size. 

We also implement a logistic regression (LR) on the featurized data to see how it compares to fully connected layers. Although LR under-performs in most cases, it still achieves good accuracy (little over 90\% in Table \ref{tab:accuracy}). This illustrates that the image featurization through the network works well enough that a simple algorithm like LR can perform well on this data. 

For activation functions in the CNN, a rectified linear unit (ReLU) is a common choice, thanks to its simplicity in gradient computation and prevention of vanishing gradient, which is common with other activation functions such as sigmoid or hyperbolic tangent. But, as seen in  Figure ~\ref{fig:flow}, clamping the activation at $0$ could potentially cause a lot of filters to be dead. Therefore, we also consider using a leaky ReLU activation with  $\alpha = 0.1$ based on the survey in \cite{leaky}. However, leaky ReLU turns out to not significantly improve the accuracy in our implementation (Table \ref{tab:accuracy}). 

\begin{figure}[ht!]
{\centering
\subfigure[Without drop-out and image augmentation, over-fitting seems to happen after about 10 epochs as the validation accuracy separates from the training accuracy.]{\includegraphics[width = 8cm]{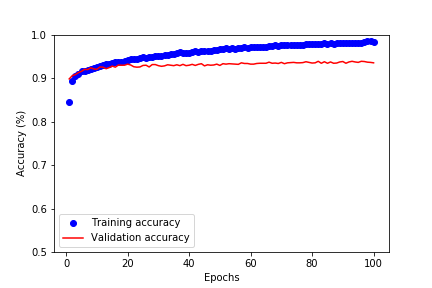}}
\subfigure[No apparent sign of over-fitting can be seen as the validation accuracy follows the training accuracy.]{\includegraphics[width = 8cm]{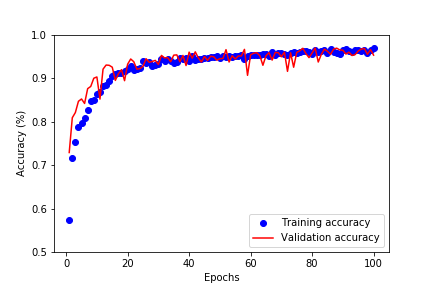}}
\\
\caption{{Over-fitting is prevented using data augmentation, drop-out, and regularization.}}
\label{fig:dropout}
}
\end{figure}

To counter over-fitting, which is a recurrent problem of deep learning, we also adopt data augmentation in the training set through random rotation, horizontal flip, vertical and horizontal shift, shear, and zoom. This can effectively increase the number of training samples to ensure better generalization and achieve better validation and test accuracy (Note that we do not perform data augmentation in the validation and test sets). Furthermore, we also employ 50\% dropout and L2 regularization with $\lambda = 10^{-6}$ in the fully connected layer. Dropout \cite{dropout} is an effective method to prevent over-fitting, especially in neural networks with many neurons. The method prevents neurons from remembering too much training data by dropping out randomly chosen neurons and their connections during the training time. %When the network is used to predict on test data, all the neurons, which were trained separately due to drop out, will act together to improve classification accuracy on unseen data. 
L2 regularization is one of the regularization techniques that has been shown to perform better on ill-poised problems or noisy data. Early application of the regularization in computer vision can be traced back to edge detection in images where the changes in intensity in an image are considered noisy \cite{Bertero}. These measures are shown to fight over-fitting effectively and significantly improve the validation accuracy in Figure~\ref{fig:dropout}.

% \begin{figure}[h]
% \hspace*{-0.4cm}{\centering
% \subfigure[Over-fitting happens after a few epochs]{\includegraphics[trim= 0 0 0 21, clip, width=3.37in]{images/4-acc-transfer.png}}
% \hspace*{0.3cm}\subfigure[Little sign of over-fitting]{\includegraphics[trim= 5 0 0 20, clip, width=2.84in]{images/2-acc-100epochs-aug-drop.png}}
% \\
% \caption{\small{Prevent over-fitting using data augmentation, drop-out, and regularization}}
% \label{fig:dropout}
% }
% \end{figure}

\begin{figure}[ht!]
{\centering
\subfigure[Transfer learning using pre-built network]{\includegraphics[width = 8cm]{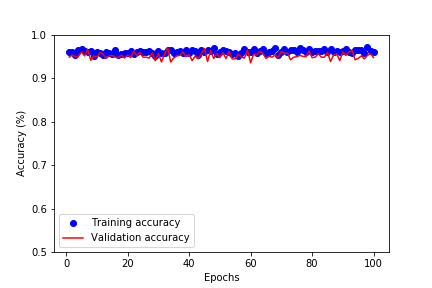}}
\subfigure[Custom network]{\includegraphics[width = 8cm]{images/no_overfit.png}}
\\
\caption{\small{Comparison between using a pre-built network and our network. The two networks almost have the same level of performance except our network achieves a slightly better accuracy with a much smaller network size. It is also noticeable that due to large number of pre-trained parameters, the bigger network achieves high accuracy right at the beginning but fails to improve subsequently. }}
\label{fig:Transfer_vs_Custom}
}
\end{figure}

As mentioned in Section ~\ref{sec:featurization}, we consider using a pre-built architecture VGG-16 (transfer learning) and building a network from scratch. In Figure~\ref{fig:Transfer_vs_Custom}, we see that the deeper and larger network can achieve a high-level validation accuracy earlier, but the accuracy pretty much plateaus (\textit{i.e.}, over-fitting happens) after a few epochs. Our simpler network can facilitate learning gradually, where the validation accuracy keeps increasing to achieve a higher value than the deeper network, and takes about 75\% less training time. 

% \begin{figure}[h]
% \hspace*{-1cm}{\centering
% \subfigure[Transfer learning using pre-built network]{\includegraphics[trim= 0 0 0 20.2, clip, width=3.676in]{images/8-everything-reulu-adam-acc.png}}
% \subfigure[Custom network]{\includegraphics[trim= 0 0 0 19, clip,width=2.85in]{images/2-acc-100epochs-aug-drop-adam.png}}
% \\
% \caption{\small{Comparison between using a pre-built network and our custom network}}
% \label{fig:Transfer_vs_Custom}
% }
% \end{figure}

We use two adaptive, momentum-based optimizers, RMSprop and Adam \cite{Adam}, with the initial learning rate of $10^{-4}$. Adam generally leads to about 1\% higher validation accuracy and less noisy learning in our implementation. %(Figure~\ref{fig:Adam_vs_RMSprop}). 

% \begin{figure}[h]
% {\centering
% \subfigure[RMSprop]{\includegraphics[trim= 4 0 0 20.5, clip,width=2.95in]{images/2-acc-100epochs-aug-drop.png}}
% \subfigure[Adam]{\includegraphics[trim= 0 0 0 19, clip,width=3.2in]{images/2-acc-100epochs-aug-drop-adam.png}}
% \\
% \caption{\small{Comparison between using RMSprop and Adam optimizers. Adam optimizer yields less variance in validation accuracy so test accuracy can be more stable.}}
% \label{fig:Adam_vs_RMSprop}
% }
% \end{figure}

% \begin{figure}[h]
% {\centering
% \subfigure[RMSprop]{\includegraphics[trim= 4 2 0 21, clip,width=2.9in]{images/2-acc-100epochs-aug-drop.png}}
% \subfigure[Adam]{\includegraphics[trim= 0 0 3 19, clip,width=3.2in]{images/2-acc-100epochs-aug-drop-adam.png}}
% \\
% \caption{\small{Comparison between using RMSprop and Adam optimizers}}
% \label{fig:Adam_vs_RMSprop}
% }
% \end{figure}

Table \ref{tab:accuracy} summarizes the performances of various models. The best performing model is our customized network with data augmentation and dropout using Adam optimizer, which can achieve 97.08\% accuracy on the unbalanced test set. The AUC metric is computed and shows a satisfying result of 99.6\% on the unbalanced test set. {\color{black}The F1 Score metric also shows a consistent choice of the best model.}

\begin{table*}[!ht]
\caption{Model performance.}
\begin{center}
\begin{tabular}{llllll}
\multicolumn{1}{c}{\bf Model}  &\multicolumn{1}{c}{\bf \begin{tabular}{@{}c@{}}Validation \\ Accuracy \end{tabular}}
  &\multicolumn{1}{c}{\bf \begin{tabular}{@{}c@{}}Test Accuracy \\ (Balanced) \end{tabular}}
   &\multicolumn{1}{c}{\bf \begin{tabular}{@{}c@{}}Test Accuracy \\ (Unbalanced) \end{tabular}}
     &\multicolumn{1}{c}{\bf \begin{tabular}{@{}c@{}}{\color{black}F1 Score} \\ \end{tabular}}

\\ \hline %\\
CNN         &95.8\%      &94.69\%  &95.47\% &{\color{black}0.9575}\\
Leaky CNN   &96.1\%       &94.79\%  &95.27\% &{\color{black}0.9558}\\
CNN + DA + DO  &97.44\%      &96.44\%  &96.56\% &{\color{black}0.9674}\\
CNN + DA + DO (Adam) &\bf{98.06\%}     &\bf{97.29\%} &\bf{97.08\%} &{\color{black}\bf{0.9723}}\\
Transfer + DO   &93.45\%      &92.8\%   &92.8\% &{\color{black}0.9304}\\
Transfer + DA + DO  &91.1\%       &88.49\%  &85.99\% &{\color{black}0.8800}\\
LR + L2     &93.55\%      &92.2\%   &91.45\% &{\color{black}0.7713}\\
{\color{black}SVM + L2}     &{\color{black}92.02\%}      &{\color{black}91.85\%}   &{\color{black}90.95\%} &{\color{black}0.7002}\\
Transfer + DA + FDO &96.5\%       &95.34\%  &95.73\% &{\color{black}0.9594}\\
\begin{tabular}{{@{}c@{}}}
   Leaky + Transfer + \\
   DA + FDO +L2
\end{tabular} &96.13\%        &95.59\% &95.68\% &{\color{black}0.9598}\\
\begin{tabular}{{@{}c@{}}}
   Leaky + Transfer + \\
   DA + FDO + L2(Adam)
\end{tabular}&97.5\%  &96.19\% &96.21\% &{\color{black}0.9643}\\
\hline
\end{tabular}
\end{center}

\textit{Legend}: CNN: Convolutional Neural Network; Leaky: Leaky ReLU activation function, else, the default is ReLU; DA: Data Augmentation; LR: Logistic Regression {\color{black}with features built by convolutional operations}; L2: L2 regularization; {\color{black}SVM: Support vector machine classifier}; (Adam): Adam optimizer, else, the default is RMSprop optimizer; DO: 50\% dropout only in the fully connected layer; FDO: Full dropout, \textit{i.e.,} 25\% dropout after every max pooling layer and 50\% in the fully connected layer; Transfer: Transfer learning using VGG-16 architecture. 
\label{tab:accuracy}
\end{table*}

\begin{figure}[ht!]
{\centering
\subfigure[AUC of balanced test set]{\includegraphics[width = 8cm]{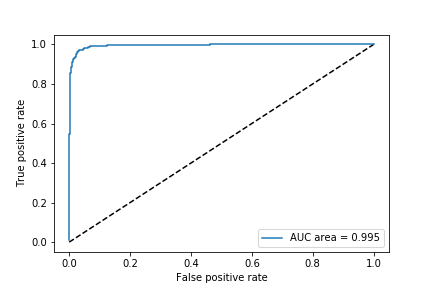}}
\subfigure[AUC of unbalanced test set]{\includegraphics[width = 8cm]{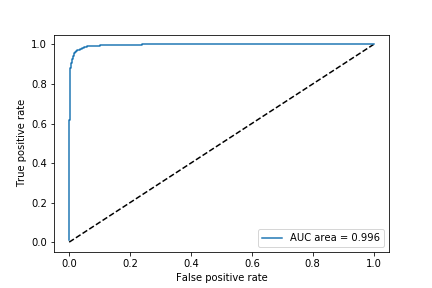}}
\\
\caption{\small{AUC for the balanced and unbalanced test sets using our best performing model---CNN + DA + DO (Adam)---in Table~\ref{tab:accuracy}.}}
\label{fig:AUC}
}
\end{figure}

\begin{figure}[ht!]
{\centering
\subfigure[]{\includegraphics[width = 3cm]{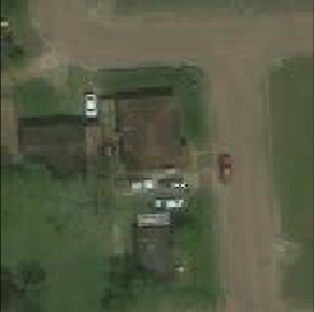}}
\subfigure[]{\includegraphics[width = 3cm]{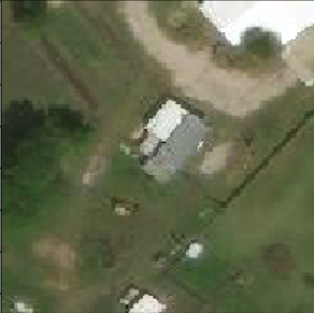}}
\subfigure[]{\includegraphics[width = 3cm]{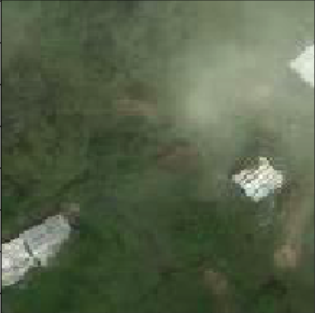}}
\subfigure[]{\includegraphics[width = 3cm]{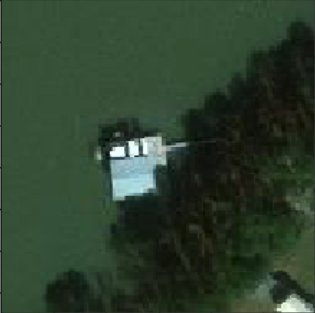}}
\subfigure[]{\includegraphics[width = 3cm]{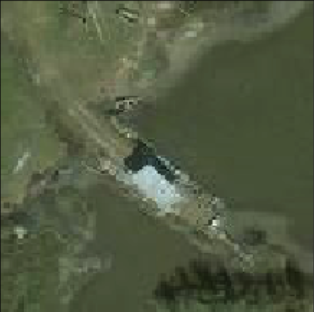}}
\subfigure[]{\includegraphics[width = 3cm]{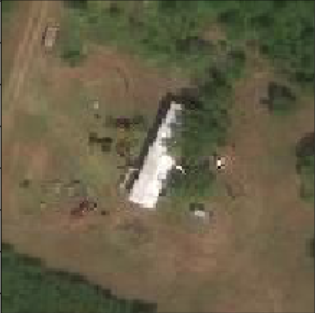}}

%\\
\caption{{False positive examples (the label is \textit{Undamaged}, whereas the prediction is \textit{Flooded/Damaged}).}}
\label{fig:false_positive}
}
\end{figure}

\begin{figure}[ht!]
{\centering
\subfigure[]{\includegraphics[width = 3cm]{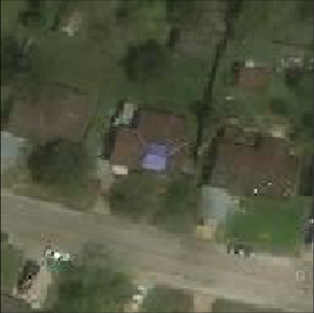}}
\subfigure[]{\includegraphics[width = 3cm]{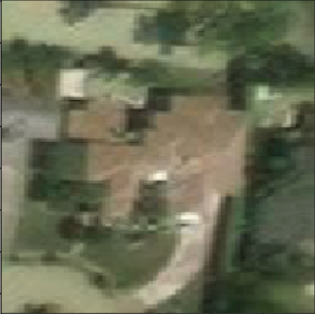}}
\subfigure[]{\includegraphics[width = 3cm]{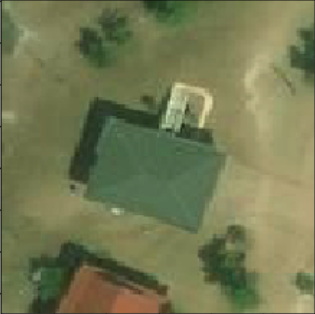}}
\subfigure[]{\includegraphics[width = 3cm]{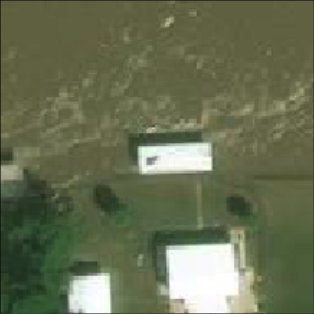}}
\subfigure[]{\includegraphics[width = 3cm]{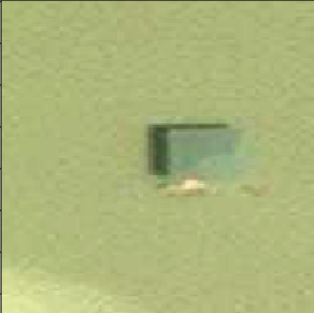}}
\subfigure[]{\includegraphics[width = 3cm]{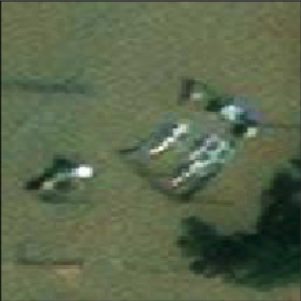}}

%\\
\caption{{False negative examples (the label is \textit{Flooded/Damaged}, whereas the prediction is \textit{Undamaged}).}}
\label{fig:false_negative}
}
\end{figure}

Although the overall result is satisfactory, we also investigate a few typical cases where the algorithm makes wrong classification to see if any intuition can be derived. Figure~\ref{fig:false_positive} shows some of the false positive cases. We hypothesize that the algorithm could predict the damage through flood water and/or debris edges. Under such hypothesis, the cars in the center of Figure~\ref{fig:false_positive}(a), the lake water in Figure~\ref{fig:false_positive}(b), the cloud covering the house in Figure~\ref{fig:false_positive}(c), and the trees covering the roof in Figure~\ref{fig:false_positive}(f) can potentially mislead the model. For the false negative cases in Figure~\ref{fig:false_negative}, it is harder to make sense out of the prediction. Even through careful visual inspection, we cannot see Figures~\ref{fig:false_negative}(a)(b) as being flooded/damaged. These could potentially be labeling mistakes by the volunteers. On the other hand, Figures~\ref{fig:false_negative}(e)(f) are clearly flooded/damaged, but the algorithm misses them. 

% \begin{figure}[h]
% {\centering
% \subfigure[]{\includegraphics[trim= 0 0 0 20, clip,width=1.5in]{images/confused_by_cars.png}}
% \subfigure[]{\includegraphics[trim= 0 0 0 20, clip,width=1.51in]{images/confused_by_roof.png}}
% \subfigure[]
% {\includegraphics[trim= 0 0 0 20, clip,width=1.52in]{images/confused_by_cloud.png}}
% \subfigure[]
% {\includegraphics[trim= 0 0 0 20, clip,width=1.55in]{images/confused_by_water.png}}
% \subfigure[]
% {\includegraphics[trim= 0 0 0 20, clip,width=1.55in]{images/confused_by_wate_roof.png}}
% \subfigure[]
% {\includegraphics[trim= 0 0 0 20, clip,width=1.5in]{images/confused_by_tree.png}}
% \\
% \caption{\small{False positive examples (label is Undamaged, prediction is Damaged).}}
% \label{fig:false_positive}
% }
% \end{figure}

% \begin{figure}[h]
% {\centering
% \subfigure[]{\raisebox{0.5mm}{\includegraphics[trim= 0 0 0 20, clip,width=1.5in]{images/volunteer_mistake.png}}}
% \subfigure[]{\includegraphics[trim= 0 0 0 20, clip,width =1.5in]{images/not_clear_damage_.png}}
% \subfigure[]
% {\includegraphics[trim= 0 0 0 20, clip,width=1.48in]{images/not_clear_damage_2.png}}
% \subfigure[]
% {\includegraphics[trim= 0 0 0 20, clip,width=1.48in]{images/not_clear_damage_3.png}}
% \subfigure[]
% {\includegraphics[trim= 0 0 0 20, clip,width=1.5in]{images/machine_mistake.png}}
% \subfigure[]
% {\includegraphics[trim= 0 0 0 21, clip,width=1.5in]{images/machine_mistakes.png}}
% \\
% \caption{\small{False negative examples (label is Damaged, prediction is Undamaged).}}
% \label{fig:false_negative}
% }
% \end{figure}

%  \begin{table}[h]
% \centering
% \caption{Comparison between the metamodel-based SIS and the EMCE for the case study}
% \label{tab:case}
% \begin{tabular}{ccccc}
% \hline
% Response & Method    & Mean   & Standard Error & CMC Ratio \\ \hline
% Edgewise & Metamodel & 0.0486 & 0.0018         & 7.0\%       \\
%          & EMCE      & 0.0486 & 0.0015         & 4.9\%       \\
% Flapwise & Metamodel & 0.0514 & 0.0028         & 32\%        \\
%          & EMCE      & 0.0535 & 0.0030         & 37\%        \\ \hline
% \end{tabular}
% \end{table}
%M/N_T = 30 % for both methods

% In the metamodel-based SIS, the metamodel is carefully built by fitting a nonhomogeneous generalized extreme value distribution to the pilot data \cite{Choe2015}. As such,  we can see that the performance of EMCE is comparable to that of metamodel-based SIS with a high quality  metamodel.  However, as seen in Section~\ref{sec-ex-sis}, when the metamodel quality is not good enough, EMCE provides a better computational efficiency. Since EMCE is an automated method, it can be a promising method when building a metamodel is difficult.    %  when  and Such a comparable performance of EMCE is promising,  whereas the metamodel-based approach requires the careful metamodel building.

%{\bf 2 Figures} (scatter \& density plot together): edge \& flap unif pilot; edge \& flap Original input, SIS-meta, SIS-EMCE density plot

%%%%%%%%%%%%%%%%%%%%%%%%%%%%%%%%%%%%%%%%%%%%%%%%%%%%%%%%%%%%%%%%%%%%%%%%%%%%%%
% Discussion
%%%%%%%%%%%%%%%%%%%%%%%%%%%%%%%%%%%%%%%%%%%%%%%%%%%%%%%%%%%%%%%%%%%%%%%%%%%%%%
\section{Conclusion and Future Research}\label{sec:conclusion}
% We demonstrated that convolutional neural networks can automatically annotate flooded/damaged buildings on post-hurricane satellite imagery with high accuracy. While our data is specific to the geographical condition and building properties in the Greater Houston area during Hurricane Harvey, the model can be further improved and generalized to other future hurricane events in other regions by collecting more positives samples from other past events and negative samples from other areas.
We demonstrated that convolutional neural networks can automatically annotate flooded/damaged buildings on post-hurricane satellite imagery with high accuracy. While our data is specific to the geographical condition and building properties in the Greater Houston area during Hurricane Harvey, the model can be further improved and generalized to other future hurricane events in other regions by collecting more positives samples from other past events and negative samples from other areas. {\color{black}From more data, we can obtain both more robust models as well as sets of hyper-parameters. As mentioned in Section 3, the cropping window size is dependent on the resolution of the available imagery (the mode of collection), as well as the typical building footprint size in the region. From our grid search on the window size, as long as the window captures the building adequately, the performance does not vary significantly so the data collection operator can still have some flexibility about the collection methods (e.g., using drones).}

For faster disaster response, a model should be able to process and annotate on low-quality images. For example, images taken right after a hurricane landfall can be covered largely by cloud. Also, image providers might not have enough time to pre-process images well due to the urgency of situation. We will investigate how a model can be made robust against such noise and distortion to reliably annotate damages. %to reduce the amount of manual processing.

% Since the positive samples are limited and valuable, we hope to further investigate how to save the samples that are partially blacked out through boundary mirror and enhance contrast to cloud covered samples. 

%Through the inspection of false positive cases, we realized there could be a link to pixel-level classification to segment different damage types and levels. For instance, we can classify flooded buildings and wrecked buildings separately. More accurate classification may be achieved through more exact shapes of different types of damage. However, this requires a massive effort to label different damage shapes and types to train the model. 

% We also wish to extend the model to the annotation of road damages and debris, which could help plan effective transportation routes of medical aids, foods, or fuels to hurricane survivors. 
{\color{black}Although the current work extracts the positive and negative samples based on different temporal information of the dataset, it would be more realistic to gather the samples from different spatial information using the same timestamp after the event happens. This would make the data closer to the actual scenario when the method is targeted to be deployed after an event happens. This direction warrants further investigation in future work.} We also wish to extend the model to the annotation of road damages and debris, which could help plan effective transportation routes of medical aids, foods, or fuels to hurricane survivors.

% Handling of black boundary by mirror extension.
% Hard negative mining
% \vspace{-.3cm}

\begin{acknowledgements}
This work was partially supported by the National Science Foundation (NSF grant CMMI-1824681). We would like to thank DigitalGlobe for data sharing through their Open Data Program. We also thank Amy Xu, Aryton Tediarjo, Daniel Colina, Dengxian Yang, Mary Barnes, Nick Monsees, Ty Good, Xiaoyan Peng, Xuejiao Li, Yu-Ting Chen, Zach McCauley, Zechariah Cheung, and Zhanlin Liu in the Disaster Data Science Lab at the University of Washington, Seattle for their help with data collection and processing.
\end{acknowledgements}
%\appendices
% \section*{Appendix: dataset and code}
%\appendix
\section*{Appendix: dataset and code}
The dataset and code used in this paper are available at the first author's Github repository  \url{https://github.com/qcao10/DamageDetection}. The dataset is also available at the IEEE DataPort (DOI: 10.21227/sdad-1e56).\\

% BibTeX users please use one of
%\bibliographystyle{spbasic}      % basic style, author-year citations
\bibliographystyle{spmpsci}      % mathematics and physical sciences
\bibliography{tomnod_BIB}   % name your BibTeX data base
\end{document}